# An m-health algorithmic approach to identify and assess physiotherapy exercises in real time

Stylianos Kandylakis[1][0000-0002-2853-0500], Christos Orfanopoulos[1][0000-0001-5687-2199], Georgios Siolas[1][0000-0002-4839-7831] and Panayiotis Tsanakas[1][0000-0002-8503-5784]

[1] School of Electrical and Computer Engineering, National Technical University of Athens, GR-15780 Greece

**Abstract.** This paper describes an efficient algorithmic approach for estimating, classifying, and evaluating human body movements during exercising while at the same time identifying divergences from the respective prescribed exercises. We apply the method to a novel mobile application that accepts camera input and performs real-time movement classification. The methodologies used to implement the app include neural networks, trigonometry, and low complexity classification algorithms. We use a dynamic programming algorithm to identify and evaluate the captured movements, while its output can indicate areas with inaccuracies. Finally, we conducted extensive experiments to fine-tune the model parameters.

## 1 Introduction

During the COVID-19 pandemic, interest from both practitioners and patients in medical applications has increased dramatically, thereby making remote monitoring and treatment a new normal. Self-data tracking techniques, namely extracting data from a person's daily lifestyle, such as exercise, walking steps, body weight, food consumed, blood pressure, cigarettes smoked, etc., are essential for mHealth services [1]. Research has shown that tracking such data is effective in itself as the procedure assists both patients and medical professionals [2]. In this paper, we deal with remote supervision of body exercises through a smart device. Until recently, most movement classification applications focused on solving the static pose classification problem, with the best-known example of yoga-pose classification [3.4]. Nowadays, especially in the physiotherapy field, there is a need to classify non-static kinetic movements, such as squats, push-ups, etc. The main purpose of this work is to extend the capabilities of previous classification approaches in order to achieve robust exercise movement analysis, by detecting inaccuracies. The key element is a novel dynamic programming approach, which will be explained and assessed in detail.

## 2      App Architecture

It is rational to assume that an exercise movement can be interpreted as a sequence of static poses. Consequently, to identify a movement we must first detect a series of poses, and then use them for subsequent movement analysis. Thus, we present the following app architecture, which consists of three main subtasks, as shown in Fig. 1.

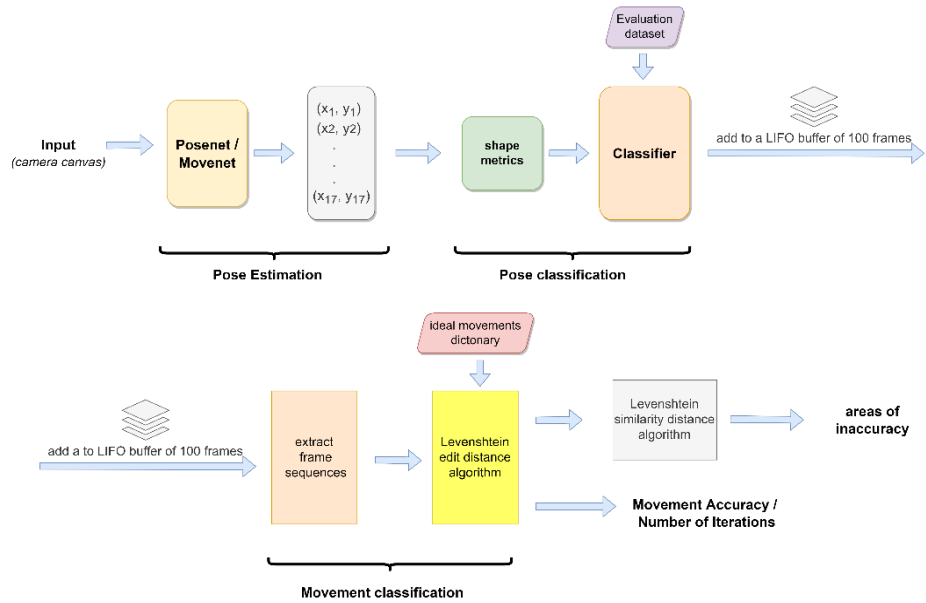

**Fig. 1.** The high-level application architecture

The first layer deals with the pose estimation. It accepts an image canvas from the camera as input. Then it processes the image through a pose estimation neural network to extract the body keypoints, stored in a JSON file.

In the second layer, the body keypoints are converted to angle-based points. Then a kNN classifier, comparing the real-time input data with an evaluation dataset, determines the performed pose and the corresponding accuracy. The results are stored in a so-called frame.

In the third stage, the previously produced frames are passed through a LIFO buffer. The frame sequences are converted to text strings. Then each string is compared via a modified Levenshtein distance algorithm to every optimal sequence of a movement dictionary. The most compatible sequence is found and then a second Levenshtein distance algorithm determines the similarity between the produced and the compatible sequence. The results of the second Levenshtein algorithm can then be used to determine the areas of inaccuracy.

It's worth mentioning that the above is a client-side architecture, thus no processing on server's side is needed [5]. The only input parameters needed are the evaluation dataset and the movements dictionary, although we can store them locally and update them occasionally. This is very crucial for the scalability of the whole application [6].

## 3    Pose Estimation

In the interest of examining the correctness of a kinetic exercise certain information must be derived from the input image. This can be achieved by using pose estimation methods. Human pose estimation is a Computer Vision task that includes detecting, associating, and tracking semantic key points of the body [7]. Every single time a frame is obtained through the camera, certain body keypoints can be extracted. Known tools to achieve this goal include classical image processing and Convolutional Neural Networks (CNN). Because image processing is a complex and time-consuming task, neural networks are preferred, especially for mobile apps. Moreover, many pre-trained models exist for that purpose. Table 1 presents the most common pose estimation libraries. PoseNet and MoveNet were chosen for the app, because of the sufficient number of keypoints being returned, 17 in total.

**Table 1.** The most commonly used pose estimation libraries

| Model | Type of estimations | Variants |
| --- | --- | --- |
| **OpenPose**[1] | Multi-person | Lightweight |
| **MoveNet**[2] | Single | Lightning / Thunder |
| **PoseNet**[3] | Single/Multi | ResNet50/ MobileNetV1 |
| **DensePose**[4] | Single/Multi | - |
| **AlphaPose (RMPE)**[5] | Single/ Multi | - |
| **BlazePose (GHUM)**[6] | Single | Heavy / Full / Light |
| **DCPose**[7] | Single/Multi | - |
| **TransPose**[8] | Single | - |
| **DeepPose**[9] | Single | - |
| **HRNet**[10] | Single | HigherHRNet / HigherHRNet+ |
| **DeepCut**[11] | Multi-person | DeeperCut |

---

[1] https://cmu-perceptual-computing-lab.github.io/openpose/web/html/doc/
[2] https://storage.googleapis.com/movenet/MoveNet.SinglePose Model Card.pdf
[3] https://arxiv.org/pdf/1505.07427.pdf
[4] http://densepose.org/
[5] https://opensourcelibs.com/lib/alphapose
[6] https://google.github.io/mediapipe/solutions/pose
[7] https://github.com/Pose-Group/DCPose
[8] https://arxiv.org/abs/2012.14214
[9] https://arxiv.org/abs/1312.4659
[10] https://arxiv.org/abs/1908.07919
[11] https://arxiv.org/abs/1511.06645

## 4   Pose Classification and Optimization

The obtained keypoints should be acquired in real time and then processed so that suitable metrics can be calculated. Keypoints are nothing more than 2D coordinates. A naive approach would be to use the extracted keypoints unaltered and add geometrical restrictions in order to classify each pose. But it is obvious that this approach is weak because evident counterexamples exist. Consequently, a more efficient solution is needed, based on trigonometry [8]. We will transform the (x,y)-keypoints to polar coordinates, so we need the following two metrics,

Euclidean distance,

$$d\big((x_1, y_1), (x_2, y_2)\big) = \sqrt{(x_1 - x_2)^2 + (y_1 - y_2)^2}, \tag{1}$$

and triangle angle,

$$angle(d_1, d_2, d_3) = \cos^{-1}\left(\frac{d_1^2 + d_2^2 - d_3^2}{2 * d_1 * d_2}\right). \tag{2}$$

As it appears in Fig. 1, the body keypoints can be mapped to designated angles in radians, creating 12 distinct body sections.

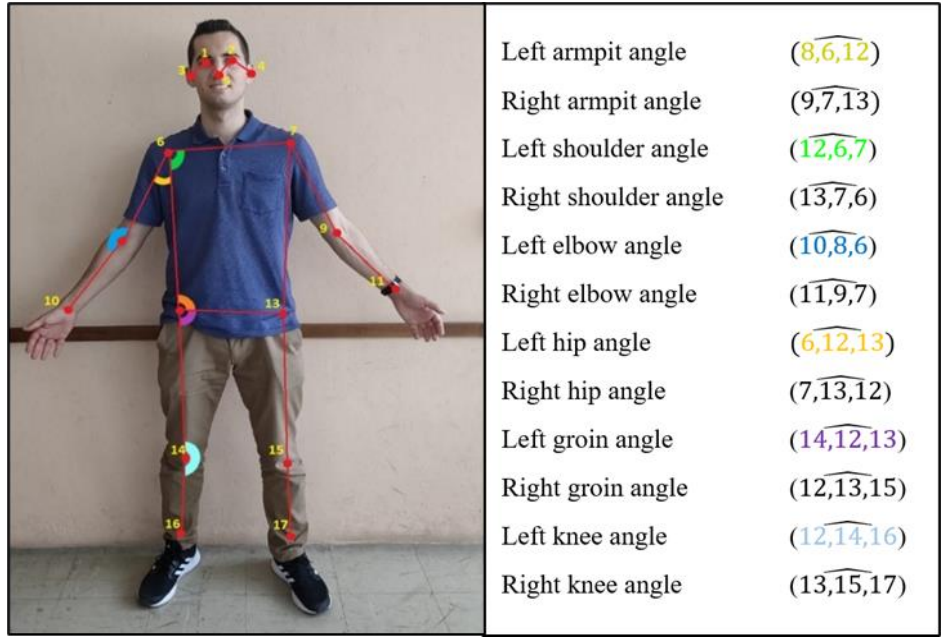

**Fig. 2.** Body keypoints and designated angles

It is important to note that these angle metrics are independent of image input resolution and body size, thus making them suitable for our app. The 12 angles correspond to the essential features mapped to each label-pose, after training the model on a dataset.

For the training process "ideal" exercise videos were used, split into frames of multiple frequencies. Each frame is passed through the neural network and certain features are extracted with the previous method. These features were then mapped to a specific pose based on our judgment. The aim is to achieve totally separated classes if possible. To optimize the model, another angle is added, called "Angular", to detect if the pose is vertical or horizontal. The angle can be seen in Fig. 3. If radians are smaller than π/4 then the pose is horizontal else, it's vertical.

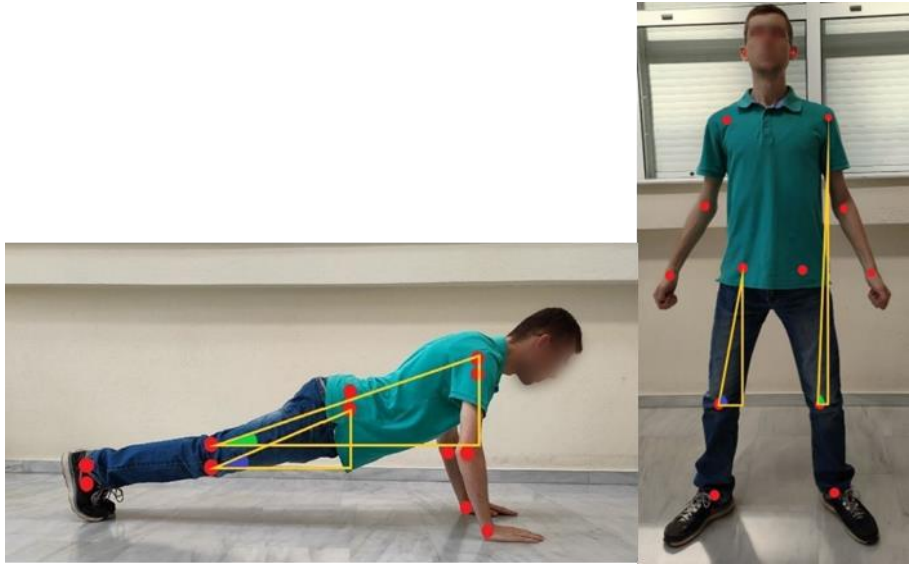

**Fig. 3.** Angular position metric

In Table 2 we can see a CSV file that emerges from the training session. A binary integer called "Position" is added, related to the "Angular", which helps separate the classes more efficiently.

**Table 2.** A typical segment of a data set used for evaluation

| Left Armpit | Right Armpit | Left Shoulder | Right Shoulder | Left Elbow | Right Elbow | Left Hip | Right Hip | Left Groin | Right Groin | Left Knee | Right Knee | Angular | Position | action |
|---|---|---|---|---|---|---|---|---|---|---|---|---|---|---|
| 51.36 | 8.39 | 53.24 | 8.12 | 137.18 | 87.67 | 175.84 | 11.2 | 172.12 | 10.9 | 112.65 | 122.17 | 0.25 | -820 | A |
| 71.19 | 19.58 | 70.18 | 20.18 | 164.53 | 103.46 | 81.42 | 106.16 | 82.43 | 107.32 | 137.17 | 102.95 | 0.35 | -820 | A |
| 45.78 | 0.64 | 45.78 | 0.64 | 164.35 | 66.07 | 120.37 | 51.18 | 121.02 | 50.08 | 134.28 | 98.01 | 0.13 | -820 | A |
| 36.97 | 21.97 | 35.23 | 21.86 | 155.64 | 42.03 | 58.54 | 113.69 | 54.52 | 112.87 | 155.6 | 102.41 | 0.15 | -820 | A |
| 45.79 | 20.11 | 43.59 | 20.32 | 145.53 | 51.94 | 84.42 | 94.85 | 85.42 | 93.15 | 139.88 | 100.78 | 0.05 | -820 | B |
| 38.94 | 37.49 | 39.04 | 37.12 | 177.15 | 165.07 | 80.11 | 99.21 | 81.04 | 98.19 | 161.59 | 179.39 | 0.23 | -820 | B |
| 32.14 | 34.71 | 35.11 | 35.21 | 164.68 | 164.65 | 92.53 | 78.19 | 91.25 | 79.15 | 163.06 | 166.16 | 0.88 | 820 | C |
| 44.61 | 37.11 | 41.67 | 40.11 | 178.96 | 179.45 | 114.36 | 80.75 | 112.78 | 83.90 | 161.06 | 166.53 | 1.01 | 820 | C |
| 42.53 | 41.07 | 41.43 | 42.84 | 166.24 | 173.97 | 78.26 | 103.57 | 79.16 | 105.13 | 161.79 | 141..02 | 0.97 | 820 | C |
| 47.22 | 41.55 | 48.76 | 41.55 | 174.16 | 177.18 | 101.72 | 87.01 | 104.25 | 85.14 | 157.89 | 135.75 | 0.91 | 820 | C |

The table features are matched to a class so they can be mapped on multi-dimensional space for supervised classification. We do dimensionality reduction (PCA) on the whole CSV evaluation file, and we can observe the results.

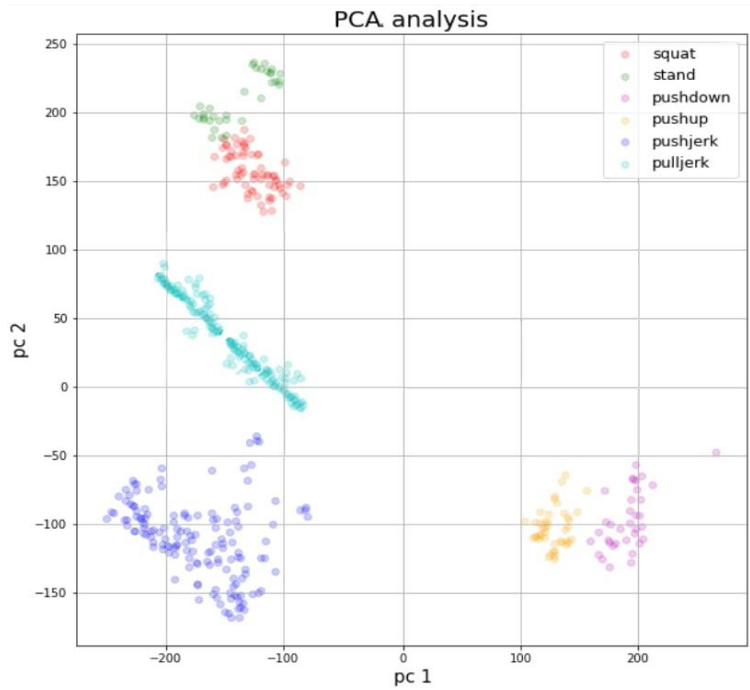

**Fig. 4.** Classification results

The mapped classes are almost linearly separated. A kNN algorithm or SVMs could classify the pose of a real-time evaluated point.

The most important evaluation metric though is the accuracy rate for the predicted pose. As shown before, features of the trained dataset can be mapped as 2D points using PCA. Specifically, given an unclassified point, the k nearest neighbors are chosen and sorted to an array. Because we are dealing with supervised learning every neighbor corresponds to a certain pose. Subsequently, the point will be classified to the most populous group of poses. The accuracy is defined as the number of same class neighbors divided by the total number of neighbors. This accuracy metric will then be used to calculate the total movement accuracy of the exercise.

## 5 Algorithm for Exercise Movements Analysis

### 5.1 Extracting the pose frame sequence

Until this point, we have analyzed how to perform static pose classification in an optimized way that can be executed on mobile devices. By static pose classification we mean that every single time a frame is obtained through the camera, we extract the body keypoints, we convert them to angle-based metrics and finally we run a classification algorithm to identify the pose.

As mentioned before a movement can be interpreted as a sequence of distinct poses. For example, we assume that a movement X is the composition of the poses A, B and C in the following sequence ABCBA (note that most physiotherapy exercises have a repetitive pattern). Thus, the goal is to identify the movement given a sequence of poses. A trivial approach would be to use a kNN classifier similar to the previous one, but this would also increase the complexity of the application. We must always consider that the app is aimed at mobile devices, therefore performance is vital. So, we need a smart algorithm that can associate the produced sequences to predefined ones. For this purpose, we apply dynamic programming.

These static poses, referred to as frames from now on, have been produced by the pose classification described in chapter 4. With the convention that physiotherapy exercises are not very fast, it is rational to set a period of 150ms per frame. Hence every 150ms the previous model produces a tuple that contains a pose class and an accuracy rate. We can also assume that a single physiotherapy exercise can last for approximately 10 seconds, thus we will focus on a LIFO buffer of 100 frames or the equivalent of 15 seconds.

Considering that every frame lasts 150ms, we can expect that the correct sequence of frames would have multiple instances of them; for example, it would be something like that: AAAAAABBBBBBCCCCCCCCCCBBBBBBBBAAAAAA. This means that the user will perform the poses A and B for almost a second, then the pose C for a little bit longer, and so on. We will transcribe the above sequence to the more concise form $A_6B_6C_{10}B_6A_6$.

We will create a dictionary containing strings of movement sequences, so we can use it later as a comparison set. These "ideal" sequences are obtained from the initial evaluation set. Obviously, in real conditions we cannot expect the user to perform the exercises with the ideal pace. Therefore, the initial sequences are stored in multiple acceptable variants. For example, the movement X (ABCBC) corresponds to $A_6B_6C_{10}B_6A_6$, $A_5B_5C_9B_5A_5$, $A_8B_8C_{14}B_8A_8$, $A_{12}B_{12}C_{20}B_{12}C_{12}$, etc. We also need to tolerate a certain deviation in the performance. For instance, we could assume that the sequence $A_5B_7C_{12}B_5A_6$ resembles the ideal $A_6B_6C_{10}B_6A_6$. As described below, we tolerate this deviation by increasing the acceptance limit for the edit distance.

Before moving on explaining the algorithm, it is important to notice that a frame can be identified as NULL, if the accuracy rate is lower than a threshold; in our case we empirically set the limit to 60%. The NULL frames are notably useful since they function as separators between sequential movements. Although attention is required to the fact that a single NULL frame is not to be considered a reliable separator. For instance, the sequence $A_6B_6C_5N_1C_4B_6A_6$, where N is NULL, should not be interpreted as two movements, but as the $A_6B_6C_{10}B_6A_6$, where a C pose was not identified correctly. Consequently, only a certain repetition of NULL frames must be considered as a separator. In our case, we use 7 NULL frames as a separator, to wit 1 second duration.

To summarize after the first stage of static pose classification the LIFO buffer of frames looks like this (table 3):

Table 3. The LIFO buffer

| No | 0 | 1 | 2 | 3 | 4 | 5 | 6 | 7 | … | … | 99 |
|---|---|---|---|---|---|---|---|---|---|---|---|
| **Class** | N | A | B | B | N | C | C | C | … | … | A |
| **Accuracy** | - | 90% | 45% | 60% | - | 96% | 93% | 66% | … | … | 83% |

### 5.2 Dynamic programming to identify movements

This buffer works as the input for the dynamic programming algorithm.

The first step is to extract the pose sequences, using the NULL separators. As a result, we obtain some strings of sequences, such as $A_6B_6C_5N_1C_4B_6A_6$. The goal is to associate this string to a predefined one. This problem is well documented and is solved by special programs known as spell checkers/correctors. The predefined strings are stored in the aforementioned dictionary, and they represent the acceptable physiotherapy exercises.

There are many implementations of spell checkers/correctors; some of them are vastly advanced [9]. Most spell correctors work by comparing the given word to a list of words in a lexicon, then finding the most suitable one by calculating the Levenshtein/edit distance [10] and finally picking the word with the minimum distance [11]. The edit distance problem is a common use case of Dynamic Programming. Furthermore, there are many complex ways to augment the performance by eliminating candidate words by using filters such as Bloom Filters [12], or by achieving faster candidate hashing [13]. Though, in our case, a simple brute-force approach, where we try all candidate strings with a given edit distance, can work as well due to the fact that the dictionary for the physiotherapy exercises is relatively small. For instance, a dictionary of 3 exercises contains about 40 words. The small size of the dictionary, even for more than 400 words, allows a brute-force approach to run in less than 150ms (namely the frame rate). We also need to exclude words with edit distance larger than a limit to avoid a high false identification rate. In our case we set that limit to 10. It is also feasible to run computations in parallel to achieve better performance. [14]. Besides, nowadays most phones have multicore/multithread CPUs.

To summarize, after extracting the frame sequence, we find the corresponding movement with the minimum edit distance. To calculate the edit distance, we use dynamic programming (Levenshtein distance algorithm). We exclude words with edit distance higher than 10. It is worth mentioning that after correctly identifying a frame sequence, we need to remove it from the buffer (set it to NULL), so it won't be examined again.

### 5.3   Locating inaccuracies and calculating mean accuracy

One of the main goals of the described application is to locate inaccuracies in exercises that the user performs, so it can later provide feedback on how to improve the therapeutic procedure. This novel idea can be implemented by comparing the produced frame sequence to the corresponding ideal one. For this purpose, we can alter the previous edit distance algorithm, so that it stores the index position, where the incompatibility was found [15,16]. We can also use the partial accuracy for each frame and spot the region with the lowest mean accuracy. This can be easily implemented by segmenting the frame sequence to shorter ones, calculating the mean average accuracy in each segment, and locating the minimum. These two methods can be used combined, and the information produced can be stored to logs for subsequent processing of multiple movements.

On the other hand, to estimate the mean accuracy for the whole frame sequence we cannot simply calculate the average accuracy of the individual frames. This may be sufficient enough to locate inaccuracies, but it won't calculate the mean accuracy correctly, considering that we should compare the produced sequence to the ideal. Thus, we will use a modified version of the Levenshtein distance algorithm to calculate the similarity score between the two.

As mentioned before the Levenshtein distance between the produced and the ideal sequence is defined as the minimum number of edits (insertions, deletions, substitutions) needed to transform the produced sequence into the ideal one [17]. Therefore, to extract a similarity percentage we divide the sum of compatible frames that do not require edit, after multiplying them with their individual accuracy rate, with the total number of ideal frames.

For instance, let's assume the following produced sequence:

**Table 4.** Example pose sequence

| No | 0 | 1 | 2 | 3 |
|---|---|---|---|---|
| **produced sequence** | A | B | C | B |
| **accuracy** | 90% | 94% | 85% | 87% |
| **ideal sequence** | A | B | B | A |

Note that 2-C is incorrect. The total accuracy is calculated as:

$$\text{TotalAcurracy} = (1*0.9 + 1*0.94 + 0*0.85 + 1*0.87) / 4 = 67.75\%$$

A similar implementation of the Levenshtein distance similarity score, can be found online [18].

An optimized algorithm could use different to zero weights for incompatible poses. Namely, if they correspond to adjacent poses, we set the weight to 0.5, else if the misidentified pose corresponds to an unfamiliar (for this movement) pose the weight is set to 0. Of course, the weight is equal to 1 when the pose is compatible.

## 6  Discussion and Conclusions

This paper presents a thorough movement accuracy-based identification application for mobile devices. The proposed application is capable of classifying, in real-time, the movements a user performs, by analyzing a series of static poses obtained from the camera of an average smartphone, and then process the same sequence to identify inaccuracies. The required processing is accomplished by deploying a dynamic programming algorithm to find the Levenshtein edit distance. So, the movement can be identified from a sequence of poses. To the best of our knowledge, this is the first integrated application that uses the Levenshtein algorithm to achieve movement estimation and examination.

The main advantages of the proposed application:

- it is suitable for physiotherapists who want to asynchronously monitor their patients
- it can be easily implemented, and prerequisite is a medium range mobile device
- it is fast and scalable, as it is client-based

Future improvements could be the classification of more complicated exercises, like the sequence of movements, instead of sequence of poses. The application could also include hand therapy exercises since it is already feasible to recognize static hand gestures [19]. Furthermore, the initial static pose can be extracted by 3D pose estimation models, thus identifying 3D body keypoints, for enhanced precision. This will allow the methodology to be extended to achieve more complicated body performance monitoring, like dancing practices, which can be covered by the forthcoming, more powerful, mobile devices.